\pdfoutput=1

\documentclass[11pt]{article}

\usepackage[]{acl}

\usepackage{times}
\usepackage{latexsym}

\usepackage[T1]{fontenc}

\usepackage[utf8]{inputenc}

\usepackage{microtype}

\usepackage{microtype}
\usepackage{graphicx}
\usepackage{url}
\usepackage{latexsym}
\usepackage{tabularx}
\usepackage{adjustbox}
\usepackage{multirow}
\usepackage{booktabs}
\usepackage{xcolor}
\usepackage{amsmath}
\usepackage{makecell}
\usepackage{hyperref}
\usepackage{nameref}
\usepackage{makecell}
\usepackage{amssymb}
\usepackage{pifont}

\newcommand\wecEng{WEC-Eng}

\definecolor{greenrgb}{rgb}{0.18, 0.71, 0.18}
\definecolor{lightblue1}{rgb}{0.6, 0.81, 0.93}
\definecolor{lightblue2}{rgb}{0.45, 0.76, 0.98}
\definecolor{lightblue3}{rgb}{0.35, 0.76, 0.98}

\newcommand{\cmark}{\ding{51}}%
\newcommand{\xmark}{\ding{55}}%

\title{WEC: Deriving a Large-scale Cross-document Event Coreference dataset from Wikipedia}

\author{Alon Eirew\textsuperscript{1,2} \quad
    Arie Cattan\textsuperscript{1} \quad
    Ido Dagan\textsuperscript{1} \\
\textsuperscript{1}Bar Ilan University, Ramat-Gan, Israel \quad \textsuperscript{2}Intel Labs, Israel \\
  {\tt  alon.eirew@intel.com} \quad {\tt  arie.cattan@gmail.com} \quad \\ {\tt  dagan@cs.biu.ac.il} \\
  }

\date{}

\begin{document}
\maketitle
\begin{abstract}
Cross-document event coreference resolution is a foundational task for NLP applications involving multi-text processing. However, existing corpora for this task are scarce and relatively small, while annotating only modest-size clusters of documents belonging to the same topic. 
To complement these resources and enhance future research, we present \textbf{W}ikipedia \textbf{E}vent \textbf{C}oreference (\textbf{WEC}), an efficient methodology for gathering a large-scale dataset for cross-document event coreference from Wikipedia, where coreference links are not restricted within predefined topics. We apply this methodology to the English Wikipedia and extract our large-scale \textbf{\wecEng{}} dataset. Notably, our dataset creation method is generic and can be applied with relatively little effort to other Wikipedia languages. To set baseline results, we develop an algorithm that adapts components of state-of-the-art models for within-document coreference resolution to the cross-document setting. Our model is suitably efficient and outperforms previously published state-of-the-art results for the task.
\end{abstract}

\section{Introduction}
\label{intro}

\emph{Cross-Document (CD) Event Coreference} resolution is the task of identifying clusters of text mentions, across multiple texts, that refer to the same event. Successful identification of such coreferring mentions is beneficial for a broad range of applications at the multi-text level, which are gaining increasing  interest and need to match and integrate information across documents, such as multi-document summarization~\cite{falke-etal-2017-concept,liao-etal-2018-abstract}, multi-hop question answering \cite{dhingra-etal-2018-neural, wang-etal-2019-multi-hop} and Knowledge Base Population (KBP)~\cite{lin2020kbpearl}. 

Unfortunately, rather few datasets of reasonable scale exist for CD event coreference. Notable datasets include ECB+~\cite{cybulska-vossen-2014-using}, MEANTIME~\cite{minard-etal-2016-meantime} and the Gun Violence Corpus (GVC)~\cite{vossen-etal-2018-dont} (described in Section~\ref{sec:related_work}), where recent work has been evaluated solely on ECB+. When addressed in a direct manner, manual CD coreference annotation is very hard due to its worst-case quadratic complexity, where each mention may need to be compared to all other mentions in all documents. Indeed, ECB+ contains less than 7000 event mentions in total (train, dev, and test sets). 
Further, effective corpora for CD event coreference are available mostly for English, limiting research opportunities for other languages.
Partly as a result of this data scarcity, rather little effort was invested in this field in recent years, compared to dramatic recent progress in modeling within-document coreference. 



Furthermore, most existing cross-document coreference datasets are restricted in their scope by two inter-related characteristics. First, these datasets annotate sets of documents, where the documents in each set all describe the \textit{same topic}, mostly a news event (consider the Malaysia Airlines crash as an example). 
While such topic-focused document sets guarantee a high density of coreferring event mentions, facilitating annotation, in practical settings the same event might be mentioned across an entire corpus, being referred to in documents of varied topics. 
Second, we interestingly observed that event mentions may be (softly) classified into two different types. One type, which we term a \textit{descriptive} mention, pertains to a mention involved in presenting the event or describing new information about it. For example, news about the Malaysian Airline crash will include mostly descriptive mentions of the event and its sub-events, such as \emph{shot-down}, \emph{crashed} and \textit{investigated}. Naturally, news documents about a topic, as in prior event coreference datasets, include mostly descriptive event mentions.
The other type, which we term a \textit{referential} mention, pertains to mentions of the event in sentences that do not focus on presenting new information about the event but rather mention it as a point of reference. For example, mentions referring to the airplane crash, such as \emph{the Malaysian plane crash}, \emph{Flight MH17} or \emph{disaster} may appear in documents about the war in Donbass or about flight safety.
Since referential event mentions are split across an entire corpus, they are less trivial to identify for coreference annotation, and are mostly missing in current news-based datasets. As we demonstrate later, these two mention types exhibit different lexical distributions and seem to require corresponding training data to be properly modeled.

In this paper, we present the \textbf{W}ikipedia \textbf{E}vent \textbf{C}oreference methodology (\textbf{WEC}), an efficient method for automatically gathering a large-scale dataset for the cross-document event coreference task. 
Our methodology effectively complements current datasets in the above-mentioned respects: data annotation is boosted by leveraging available information in Wikipedia, practically applicable for any Wikipedia language; mentions are gathered across the entire Wikipedia corpus, yielding a dataset that is not partitioned by topics; and finally, our dataset consists mostly of referential event mentions.

\begin{figure}[!t]
\centering
\includegraphics[width=.45\textwidth]{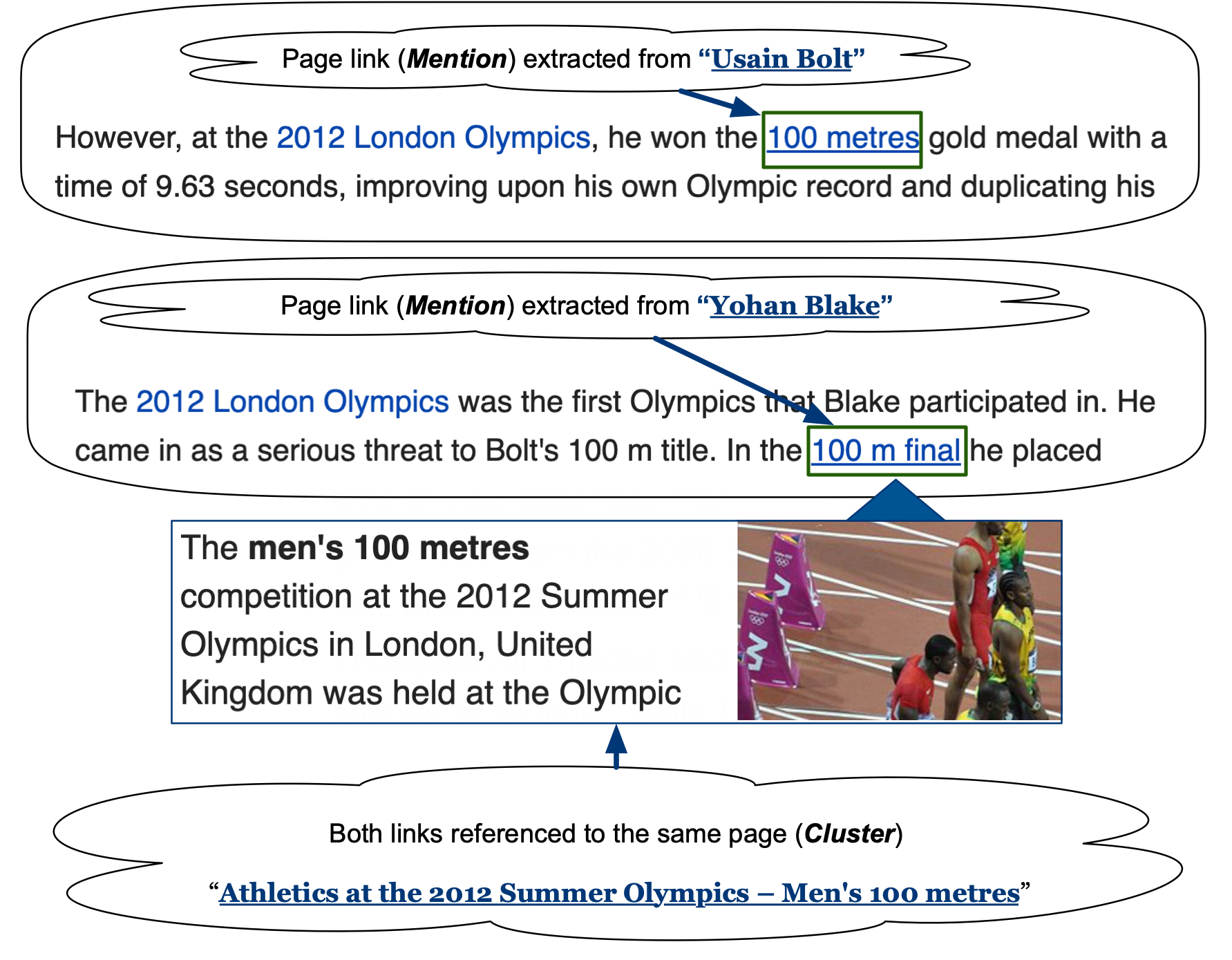}
\caption{Example of two anchor texts (\emph{100 metres}, \emph{100 m final}) from different Wikipedia articles (Usain Bolt, Yohan Blake) pointing to the same event.}
\label{fig:figure1}
\end{figure}

In its essence, our methodology leverages the coreference relation that often holds between anchor texts of hyperlinks pointing to the same Wikipedia article (see Figure \ref{fig:figure1}), similar to the basic idea introduced in the Wikilinks dataset \cite{singh12:wiki-links}.
Focusing on CD \emph{event} coreference, we identify and target only Wikipedia articles denoting events. 
Anchor texts pointing to the same event article, along with some surrounding context, become candidate mentions for a corresponding event coreference cluster, undergoing extensive filtering.
We apply our method to the English Wikipedia and extract \textbf{\wecEng{}}, our English version of a WEC dataset.
The automatically-extracted data that we collected provides a training set of a very large scale compared to prior work, 
while our development and test sets underwent relatively fast manual validation.


Due to the large scale of the \wecEng{} training data, current state-of-the-art CD coreference models cannot be easily trained and evaluated on it, for scalability reasons. We therefore developed a new, more scalable, baseline model for the task, while adapting components of recent competitive within-document coreference models \cite{lee-etal-2017-end,kantor-globerson-2019-coreference,joshi-etal-2019-bert}. In addition to setting baseline results for \wecEng{}, we assess our model's competitiveness by presenting a new state-of-the-art on the commonly used ECB+ dataset.
Finally, we propose that our automatic extraction and manual validation methods may be applied to generate additional annotated datasets, particularly for other languages.
Overall, we suggest that future cross-document coreference models should be evaluated also on the \wecEng{} dataset, and address its complementary characteristics, while the WEC methodology may be efficiently applied to create additional datasets. To that end, our dataset and code\footnote{WEC--\url{https://github.com/AlonEirew/extract-wec}}\footnote{Model--\url{https://github.com/AlonEirew/cross-doc-event-coref}} are released for open access.

\begin{table*}
    \centering
    \resizebox{.7\textwidth}{!}{
    \begin{tabular}{@{}lcccccc@{}}
        \toprule
        & \makecell{Topics} & \makecell{Mentions} & \makecell{Clusters} &  \makecell{Non-Singleton \\ Clusters} & \makecell{Ambiguity} & \makecell{Diversity} \\
        \midrule
        \wecEng{} (train) & - & 40,529 & 7,042 & 6,210 & 6.3 & 2.1 \\
        \wecEng{} (dev) & - & 1,250 & 233 & 216 & 2.7 & 1.7 \\
        \wecEng{} (test) & - & 1,893 & 322 & 306 & 2.6 & 1.9 \\
        \midrule
        ECB+ (train) & 25 & 3,808 & 1,527 & 411 & 1.4 & 2.2 \\
        ECB+ (dev) & 8 & 1,245 & 409 & 129 & 1.4 & 2.3 \\
        ECB+ (test) & 10 & 1,780 & 805 & 182 & 1.4 & 2.2 \\
        \midrule
        GVC (all) & 1 & 7,298 & 1,411 & 1,046 & 19.5 & 3.0 \\
        \midrule
        MEANTIME (all) & 4 & 2,107 & 1,892 & 142 & 2.1 & 1.5 \\
        \bottomrule
    \end{tabular}}
    \caption{Event Coreference Datasets Statistics (No train/dev/test split exists for MEANTIME and GVC). \textbf{Non-Singleton Clusters}: Number of clusters with more than a single mention.  \textbf{Ambiguity}: Average number of different clusters in which a head lemma appears. \textbf{Diversity}: Average number of unique head lemmas within a cluster (excluding singletons for fair comparison).}
    \label{tab:table_datasets_stats}
\end{table*}

\section{Related Datasets}
\label{sec:related_work}

This section describes the main characteristics of notable datasets for CD event coreference (ECB+, MEANTIME, GVC).
Table~\ref{tab:table_datasets_stats} presents statistics for all these datasets, as well as ours. 
We further refer to the Wikilinks dataset, which also leveraged Wikipedia links for CD coreference detection.

\subsection{CD Event Corpora}
\paragraph{ECB+} This dataset \cite{cybulska-vossen-2014-using}, which is an extended version of the EventCorefBank (ECB) \cite{bejan-harabagiu-2010-unsupervised}, is the most commonly used dataset for training and testing models for CD event coreference \cite{choubey-huang-2017-event,kenyon-dean-etal-2018-resolving,barhom-etal-2019-revisiting}. This corpus consists of documents partitioned into 43 clusters, each corresponding to a certain news topic. In order to introduce some ambiguity and to limit the use of lexical features, each topic is composed of documents describing two different events (called \emph{sub-topics}) of the same event type (e.g. two different celebrities checking into rehab facilities). Nonetheless, as can be seen in Table~\ref{tab:table_datasets_stats}, the ambiguity level obtained is still rather low.
ECB+ is relatively small, where on average only 1.9 sentences per document were selected for annotation, yielding only 722 non-singleton coreference clusters in total.

\paragraph{MEANTIME} \newcite{minard-etal-2016-meantime} proposed a dataset that is similar in some respects to ECB+, with documents partitioned into a set of topics. 
The different topics do not correspond to a specific news event but rather to a broad topic of interest (e.g. Apple, stock market). Consequently, different documents rarely share coreferring event mentions, resulting in only 11 event coreference clusters that include mentions from multiple documents, making this dataset less relevant for training CD coreference models. 

\paragraph{Gun Violence Corpus (GVC)} 
This dataset \cite{vossen-etal-2018-dont} was triggered by the same motivation that drove us, of overcoming the huge complexity of direct manual annotation of CD event coreference from scratch. 
To create the dataset, the authors leveraged a structured database recording gun violence events, in which the record for an individual event points at documents describing that event. The annotators were then asked to examine the linked documents and mark in them mentions of 5 gun-violence event classes (firing a gun, missing, hitting, injuring, death). Considering the recorded event as a pivot, all mentions found for a particular class were considered as coreferring. 
Using this process, they report an annotation rate of about 190 mentions per hour. 
As this corpus assumes a specific event structure scheme related to gun violence, it is more suitable for studying event coreference within a narrow domain rather than for investigating models for broad coverage  event coreference. 

\subsection{Wikilinks}
Wikilinks \cite{singh12:wiki-links} is an automatically-collected large-scale cross-document coreference dataset, focused on entity coreference.
It was constructed by crawling a large portion of the web and collecting as mentions hyperlinks pointing to Wikipedia articles. 
Since their method does not include mention distillation or validation, it was mostly used for training models for the Entity Linking task, particularly in noisy texts \cite{chisholm-hachey-2015-entity,eshel-etal-2017-named}.

\begin{table*}
    \centering
    \resizebox{1\textwidth}{!}{
    \begin{tabular}{l||l}\toprule
    \multicolumn{1}{c}{\textbf{Cluster-1} (2010 Polish Air Force Tu-154 crash)} &
    \multicolumn{1}{c}{\textbf{Cluster-2} (Lokomotiv Yaroslavl plane crash)} \\
    \toprule
    ...following the death of President Lech Kaczyński in \emph{\textcolor{blue}{a plane crash}}... & ...On 7 September 2011, nearly the entire team perished in \emph{\textcolor{blue}{an airplane crash}}... \\
    \midrule
    ...following the \emph{\textcolor{blue}{Smolensk air disaster}} which killed the incumbent Polish president... & ...fourth season was overshadowed by the \emph{\textcolor{blue}{Yaroslavl air disaster}} on 7 September... \\
    \midrule
    ...died when the \emph{\textcolor{blue}{presidential plane went down}} about a half mile from the runway... & ...Early in September, \emph{\textcolor{blue}{tragedy}} rocked the hockey world... \\
    \bottomrule
    \end{tabular}}
    \caption{Mentions from two generated \wecEng{} clusters, illustrating typical challenging ambiguity. To resolve, the model must consider the surrounding paragraph extracted along with the mention (not fully shown here).}
    \label{tab:table_clust_examp}
\end{table*}

\section{The WEC Methodology and Dataset}
\label{sec:wec}

We now describe our methodology for gathering a CD event coreference dataset from Wikipedia, and the \wecEng{} dataset created by applying it to the English Wikipedia. We also denote how this methodology can be applied, with some language-specific adjustments, to other Wikipedia languages.


\subsection{Dataset Structure}
Our data is collected by clustering together anchor texts of (internal) Wikipedia links pointing to the same Wikipedia concept. 
This is generally justified since all these links refer to the same real world theme described by that article, as illustrated in Figure~\ref{fig:figure1}. Accordingly, our dataset consists of a set of mentions, each including the mention span corresponding to the link anchor text, the surrounding context, and the mention cluster ID. Since Wikipedia is not partitioned into predefined topics, mentions can corefer across the entire corpus (unlike most prior datasets).

Since mention annotation is not exhaustive, coreference resolution is performed over the gold mentions.
Thus, our goal is to support the development of CD event coreference algorithms, rather than of mention extraction algorithms. 
Our dataset also includes metadata information, such as source and target URLs for the links, but these are not part of the data to be considered by algorithms, as our goal in this work is CD coreference development rather than Event Linking \cite{nothman2012event}.

\begin{table*}
    \centering
    \resizebox{\textwidth}{!}{
    \begin{tabular}{@{}c|l|ll@{}}
    \toprule
    \textbf{Cluster} & \multicolumn{1}{c}{\textbf{Mention link and context}} & \multicolumn{2}{c}{\textbf{Validation}}  \\
    \toprule
        \multirow{2}{*}{1} & ..The Tar Heels made it to the final of the \emph{\textcolor{blue}{2001 ACC Men's Basketball Tournamen}}.. & \textbf{\textcolor{greenrgb}{\cmark}} & \\
        & ..Duke finished the regular season with a 26–4 record entering the  \emph{\textcolor{blue}{ACC Tournamen}}.. & \textbf{\textcolor{red}{\xmark}} & Insufficient mention context \\
        \midrule
        \multirow{2}{*}{2} & ..was nominated for an Academy Award at the  \emph{\textcolor{blue}{1939 Oscars}} for Best.. & \textbf{\textcolor{greenrgb}{\cmark}} & \\
        & ..\emph{\textcolor{blue}{In the following year}}, the Academy presented Disney an Honorary Academy Award.. & \textbf{\textcolor{red}{\xmark}} & Link placed on event time \\
        \midrule
        \multirow{2}{*}{3} & ..was kidnapped during the \emph{\textcolor{blue}{Draper's Meadow massacre}} in July 1755.. & \textbf{\textcolor{greenrgb}{\cmark}} & \\
        & ..with a terrible attack on \emph{\textcolor{blue}{Drapers Meadow}} by the Shawnee Indian Warriors.. & \textbf{\textcolor{red}{\xmark}} & Link placed on event location \\
        \midrule
        \multirow{2}{*}{4} & ..War and was on board the USS ''Forrestal'' in 1967 when a \emph{\textcolor{blue}{fatal fire}} broke out.. & \textbf{\textcolor{greenrgb}{\cmark}} & \\
        & ..the repairs to the supercarrier which had been extensively \emph{\textcolor{blue}{damaged}} off Vietnam.. & \textbf{\textcolor{red}{\xmark}} & Subevent \\
    \bottomrule
    \end{tabular}}
    \caption{Examples of \wecEng{} manually validated mentions, showing validation results (validated/disqualified)}
    \label{tab:validation}
\end{table*}

\subsection{Data Collection Process}
\label{auto-data-collect}

In this paper, we focus on deriving from Wikipedia an \textit{event} coreference dataset. The choice to focus on event coreference was motivated by two observations: (1) coreference resolution for Wikipedia anchor texts would be more challenging for event mentions than for entity mentions, since the former exhibits much higher degrees of both ambiguity and lexical diversity, and (2) event structures, with their arguments (such as participants, location and time) available in the surrounding context, would facilitate a more natural dataset for the corpus-wide CD coreference task, compared to Wikipedia entity mentions which are comprised mostly of named entities.

Accordingly, we seek to consider only Wikipedia pages denoting events, then collect hyperlinks pointing at these pages. All anchor texts pointing to the same event then become the mentions of a corresponding event coreference cluster, 
and are extracted along with their surrounding paragraph as context (see  Table~\ref{tab:table_clust_examp}). 
The following paragraphs describe this process in detail, and how it was applied to generate the WEC-Eng dataset from English Wikipedia.

\paragraph{Event Identification} Many Wikipedia articles contain an infobox\footnote{\url{https://en.wikipedia.org/wiki/Help:Infobox}} element. This element can be selected by a Wikipedia author from a pre-defined list of possible infobox types (e.g. “Civilian Attack”, “Game”, “Scientist”, etc.), each capturing typical information fields for that type of articles. For example, the “Scientist” infobox type consists of fields such as “birth date”, “awards”, “thesis” etc. We leverage the infobox element and its parameters in order to identify articles describing events (e.g. accident, disaster, conflict, ceremony, etc.) rather than entities (e.g. a person, organization, etc.).

To that end, we start by automatically compiling a list of all Wikipedia infobox types that are associated with at least dozens of Wikipedia articles. Of those, we manually identify all infobox types related to events (\wecEng{} examples include Awards, Meetings, Civilian Attack, Earthquake, Contest, Concert and more). We then (manually) exclude infobox types that are frequently linked from related but non-coreferring mentions, such as sub-events or event characteristics, like location and time
(see Appendix~\ref{app:subevent} for further details).
For \wecEng{}, we ended up with 28 English Wikipedia event infobox types (see Appendix~\ref{app:infobox} for the full list).

\paragraph{Gathering Initial Dataset} Once the infobox event list is determined, we apply a fully automatic pipeline to obtain an initial crude version of the dataset. This pipeline consists of: (1) Collecting all Wikipedia articles (event “pivot” pages) whose Infobox type is in our list. (2) Collecting all Wikipedia anchor texts (“mentions”) pointing to one of the pivot pages, along with their surrounding paragraph. (3) Filtering mentions that lack context or those belonging to Wikipedia metadata, such as tables, images, lists, etc., as well as mentions whose surrounding context contains obvious Wikipedia boilerplate code (i.e. HTML and JSON tags) (4) Finally, all collected mentions are clustered according the pivot page at which they point.

\paragraph{Mention-level Filtering}
\label{para:filtering}

An event coreference dataset mined this way may still require some refinement in order to further clean the dataset at the individual mention level. Indeed, we observed that many Wikipedia editors have a tendency to position event hyperlinks on an event argument, such as a Named Entity (NE) related to the event date or location (as in the case of the disqualified mention for cluster 3 in Table~\ref{tab:validation}). To automatically filter out many of the cases where the hyperlink is placed on an event argument instead of on the event mention itself, we use a Named Entity tagger and filter out mentions identified by one of the following labels: PERSON, GPE, LOC, DATE and NORP (for \wecEng{} we used the SpaCy Named Entity tagger~\cite{spacy2}). 

\paragraph{Controlling Lexical Diversity}
\label{para:diversity}
So far, we addressed the need to avoid having invalid mentions in a cluster, which do not actually refer to the linked pivot event.

Next, we would like to ensure a reasonably balanced lexical distribution of the mentions within each cluster. Ideally, it would be desired to preserve the ``natural” data distributions as much as possible. 
However, we observed that in many Wikipedia hyperlinks, the anchor texts used in an event mention may be lexically unbalanced. Indeed, Wikipedia authors seem to have a strong bias to use the pivot article title when creating hyperlinks pointing at that article, while additional ways by which the event can be referred are less frequently hyperlinked.
Consequently, preserving the original distribution of hyperlink terms would create a too low level of lexical diversity. As a result, training a model on such a dataset might overfit to identifying only the most common mention phrases, leaving little room for identifying the less frequent ones. To avoid this, we applied a simple filter\footnote{We also release the non controlled version of the dataset.} that allows a maximum of 4 mentions having identical strings in a given cluster. This hyperparameter was tuned by making the lexical repetition level in our clusters more similar to that of ECB+, in which lexical diversity was not controlled (resulting with an average of 1.9 same-string mentions per cluster in \wecEng{} train set compared to 1.3 in ECB+).

Using this process we automatically generated a large dataset. We designated the majority of the automatically generated data to serve as the \wecEng{} training set. The remaining data was left for the development and test sets, which underwent a manual validation phase, as described next.


\subsection{Manual Validation}
\label{subsec:validation}
Inevitably, some noise would still exist in the automatically derived dataset just described. While partially noisy training data can be effective, as we show later, and is legitimate to use, the development set, and particularly the test set, should be of high quality to allow for proper evaluation. To that end, we manually validated the mentions in the development and test sets.

For CD coreference evaluation, we expect to include a mention as part of a coreferring cluster only if it is clear, at least from reading the given surrounding context, that this mention indeed refers to the linked pivot event. Otherwise, we cannot expect a system to properly detect coreference for that mention. Such cases occasionally occur in Wikipedia, where identifying context is missing while relying on the provided hyperlink (see cluster-1 in Table~\ref{tab:validation}, where the tournament year is not mentioned). Such mentions are filtered out by the manual validation. Additionally, misplaced mention boundaries that do not include the correct event trigger (Table~\ref{tab:validation} cluster-2,3), as well as mentions of subevents of the linked event (Table~\ref{tab:validation} cluster-4), are filtered.

Summing up, to filter out these cases, we used a strict and easy-to-judge manual validation criterion, where a mention is considered valid only if: (1) the mention boundaries contain the event trigger phrase; (2) the mention’s surrounding paragraph suffices to verify that this mention refers to the pivot page and thus belongs to its coreference cluster; and (3) the mention does not represent a subevent of the referenced event. Table~\ref{tab:validation} shows examples of validated vs. disqualified mentions judged for the \wecEng{} development set.

For the \wecEng{} validation, we randomly selected 588 clusters and validated them, yielding 1,250 and 1,893 mentions for the development and test sets, respectively. Table~\ref{tab:table_datasets_stats} presents further statistics for \wecEng{}. The validation was performed by a competent native English speaker, to whom we explained the guidelines, after making a practice session over 150 mentions. Finally, all training mentions that appeared in the same (source) article with a validated mention were discarded from the training set.

Our manual validation method is much faster and cheaper compared to a full manual coreference annotation process, where annotators would need to identify and compare all mentions across all documents. In practice, the average annotation rate for \wecEng{} yielded 350 valid mentions per hour, 
with the entire process taking only 9 hours to complete.
In addition, since our validation approach is quite simple and does not require linguistic expertise, the eventual data quality is likely to be high. To assess the validation quality, one of the authors validated 50 coreference clusters (311 mentions), randomly selected from the development and test sets, and then carefully consolidated these annotations with the original validation judgements by the annotator. Relative to this reliable consolidated annotation, the original annotations scored at 0.95 Precision and 0.96 Recall, indicating the high quality of our validated dataset (the Cohen's Kappa  \cite{doi:10.1177/001316446002000104} between the original and consolidated annotations was 0.75, considered substantial agreement).


In all, 83\% of the candidate mentions were positively validated in the development and test sets, indicating a rough estimation of the noise level in the training set. That being said, we note that a majority of these noisy mentions were not totally wrong mentions but rather were filtered out due to the absence of substantial surrounding context or the misplacement of mention boundaries (see examples in Table~\ref{tab:validation}).

\subsection{Dataset Content}
\label{subseb:content}
This section describes the \wecEng{} dataset content and some of its characteristics. The final \wecEng{} dataset statistics are presented in Table~\ref{tab:table_datasets_stats}.
Notably, the training set includes 40,529 mentions distributed into 7,042 coreference clusters, facilitating the training of deep learning models.

The relatively high level of lexical ambiguity shown in the table\footnote{The higher ambiguity level of the training sets stems from its larger size --- including many more coreference clusters in total, and accordingly more clusters under individual event types.} is an inherent characteristic caused by many events (coreference clusters) sharing the same event type, and thus sharing the same terms, as illustrated in Table~\ref{tab:table_clust_examp}.
Identifying that identical or lexically similar mentions refer to different events is one of the major challenges for CD coreference resolution, particularly in the corpus-wide setting, where different documents might refer to similar yet different events. 


With respect to the distinction between \textit{descriptive} and \textit{referential} event mentions, proposed in Section \ref{intro}, \wecEng{} mentions are predominantly referential. This stems from the fact that its mentions correspond to hyperlinks that point at a different Wikipedia article, describing the event, while the mention's article is describing a different topic. 
On the other hand, ECB+, being a news dataset, is expected to include predominantly descriptive mentions. 
Indeed, manually analyzing a sample of 30 mentions from each dataset, in \wecEng{} 26 were referential while in ECB+ 28 were descriptive.
This difference also imposes different lexical distributions for mentions in the two datasets, as sampled in Appendix~\ref{app:verb_dist}. When describing an event, verbs are more frequently used as event mentions, but nominal mentions are abundant as well. This is apparent for the predominately descriptive  ECB+, where 62\% of the mentions in our sample are verbal vs. 38\% nominal. On the other hand, when a previously known event is only referenced, it is mostly referred by a nominal mention. Indeed, in the predominantly referential \wecEng{}, a vast majority of the mentions are nominal (93\% in our sample).



\subsection{Potential Language Adaptation}
While our process was applied to the English Wikipedia, it can be adapted with relatively few adjustments and resources to other languages for which a large-scale Wikipedia exists. Here we summarize the steps needed to apply our dataset creation methodology to other Wikipedia languages.
To generate a dataset, the first step consists of manually deciding on the list of suitable infobox types corresponding to (non-noisy) event types. Then, the automatic corpus creation process can be applied for this list, which takes only a few hours to run on a single CPU. After the initial dataset was created, a language specific named-entity tagger should be used to filter mentions of certain types, like time and location (see Mention-level Filtering (\ref{para:filtering})). Next, the criterion for ensuring balanced lexical diversity in a cluster (see Controlling  Lexical  Diversity (\ref{para:diversity})), which was based on a simple same-string test for English, may need to be adjusted for languages requiring a morphological analyzer.
Finally, as we perform manual validation of the development and test sets, this process should be performed for any new dataset (see Section~\ref{subsec:validation}). Supporting this step, our validation guidelines are brief and simple, and are not language specific. They only require identifying subevents and misplaced mention boundaries, as well as validating the sufficiency of the mention context.





\begin{table*}
    \centering
    \resizebox{\textwidth}{!}{
    \begin{tabular}{@{}lcccccccccccccc@{}}\toprule
    & \phantom{abc}&\multicolumn{3}{c}{MUC} & \phantom{abc}& \multicolumn{3}{c}{$B^3$} & \phantom{abc}& \multicolumn{3}{c}{$CEAF$} & \phantom{abc}& CoNLL\\
    \cmidrule{3-5} \cmidrule{7-9} \cmidrule{11-13} \cmidrule{15-15}
    && R & P & $F_1$ && R & P & $F_1$ && R &P & $F_1$ && $F_1$  \\ 
    \midrule
        Lemma Baseline && 76.5 & 79.9 & 78.1 && 71.7 & 85 & 77.8 && 75.5 & 71.7 & 73.6 && 76.5 \\
        \textsc{Disjoint} \cite{barhom-etal-2019-revisiting} && 75.5 & 83.6 & 79.4 && 75.4 & 86 & 80.4 && 80.3 & 71.9 & 75.9 && 78.5 \\
        \textsc{Joint} \cite{barhom-etal-2019-revisiting} && 77.6 & 84.5 & 80.9 && 76.1 & 85.1 & 80.3 && 81 & 73.8 & 77.3 && 79.5 \\
        \midrule
        Our model && 84.2 & 81.8 & 83 && 80.8 & 81.7 & 81.3 && 76.7 & 79.5 & 78 && \textbf{80.8} \\
    \bottomrule
    \end{tabular}}
    \caption{Event coreference results on the ECB+ test set}
    \label{tab:ecb-results}
\end{table*}

\begin{table*}
    \centering
    \resizebox{\textwidth}{!}{
    \begin{tabular}{@{}lcccccccccccccc@{}}\toprule
    & \phantom{abc}&\multicolumn{3}{c}{MUC} & \phantom{abc}& \multicolumn{3}{c}{$B^3$} & \phantom{abc}& \multicolumn{3}{c}{$CEAF$} & \phantom{abc}& CoNLL\\
    \cmidrule{3-5} \cmidrule{7-9} \cmidrule{11-13} \cmidrule{15-15}
    && R & P & $F_1$ && R & P & $F_1$ && R &P & $F_1$ && $F_1$  \\ 
    \midrule
        Lemma Baseline && 85.5 & 79.9 & 82.6 && 74.5 & 32.8 & 45.5 && 25.9 & 39.4 & 31.2 && 53.1 \\
       Our model && 78 & 83.6 & 80.7 && 66.1 & 55.3 & 60.2 && 53.4 & 40.3 & 45.9 && \textbf{62.3} \\
    \bottomrule
    \end{tabular}}
    \caption{Event coreference results on the \wecEng{} test set}
    \label{tab:wec-results}
\end{table*}

\section{Baseline and Analysis} 
\label{model_analysis}


\subsection{Model}


The current state-of-the-art CD event coreference system \cite{barhom-etal-2019-revisiting} cannot be effectively trained on \wecEng{} for two main reasons: (1) computational complexity and (2) reliance on verbal SRL features. With respect to computation time, the training phase of this model simulates the clustering operations done at inference time, while recalculating new mention representations and pairwise scores after each cluster merging step. Consequently, training this model on our large scale training data, which is further not segmented to topics, is computationally infeasible. In addition, the model of \newcite{barhom-etal-2019-revisiting} uses an SRL system to encode the context surrounding \emph{verbal} event mentions, while \wecEng{} is mostly composed of \emph{nominal} event mentions (Section~\ref{subseb:content}).

We therefore develop our own, more scalable, model for CD event coreference resolution, establishing baseline results for \wecEng{}.
As common in CD coreference resolution, 
we train a pairwise scorer $s(i, j)$ indicating the likelihood that two mentions $i$ and $j$ in the dataset are coreferring, and then apply agglomerative clustering over these scores to find the coreference clusters. Following the commonly used average-link method \cite{choubey-huang-2017-event,kenyon-dean-etal-2018-resolving,barhom-etal-2019-revisiting}, the merging score for two clusters is defined as the average mention pair score $s(i, j)$ over all mention pairs $(i, j)$ across the two candidate clusters to be merged.


For the pairwise model, we replicate the architecture of mention representation and pairwise scorer from the end-to-end within document coreference model in \cite{lee-etal-2017-end}, while including the recent incorporation of transformer-based encoders \cite{joshi-etal-2019-bert,kantor-globerson-2019-coreference}.
Concretely, we first apply a pre-trained RoBERTa \cite{liu2019roberta} language model (without fine-tuning), separately for each mention. Given a mention span $i$, we include as context, $T$ (set to 250) tokens to the left of $i$ and $T$ tokens to the right of $i$. 
Applying RoBERTa to this window, we represent each mention by a vector $g_i$, which is the concatenation of three vectors: the contextualized representations of the mention span boundaries (first and last) and the weighted sum of the mention token vectors according to the head-finding attention mechanism in \cite{lee-etal-2017-end}.
The two mention representations $g_i$ and $g_j$, and the element-wise multiplication of these vectors are then concatenated and fed into a simple MLP, which outputs a score $s(i, j)$, indicating the likelihood that mentions $i$ and $j$ belong to the same cluster.
The head-attention layer and the MLP are trained to optimize the standard binary cross-entropy loss over all pairs of mentions, where the label is 1 if they belong to the same coreference cluster
and 0 otherwise.\footnote{We note that this optimization is different than the (linear) antecedent ranking in the model of \newcite{lee-etal-2017-end}, since in the CD setting there is no linear order between mentions from different documents.}


\subsection{Experiments}
\label{experiments}
We first train and evaluate our model on the commonly used dataset ECB+, to assess its relevance as an effective baseline model, and then evaluate it on \wecEng{}, setting baseline results for our dataset.
We also present the performance of the challenging \emph{same-head-lemma} baseline, which clusters mentions sharing the same syntactic-head lemma.
For the experiment on ECB+, we follow the recent evaluation setting \cite{kenyon-dean-etal-2018-resolving,barhom-etal-2019-revisiting}, clustering gold mentions and concatenating all test documents into one meta-document, as proposed by \newcite{upadhyay-etal-2016-revisiting}. For a fair comparison, we use the output of pre-processing document clustering obtained by~\newcite{barhom-etal-2019-revisiting} and apply our coreference model separately on each predicted document cluster.

For both datasets, the positive examples for training consist of all the mention pairs in the dataset that belong to the same coreference cluster. 
For the ECB+ model, we consider only negative examples that belong to the same subtopic, while for \wecEng{} we sample $k$ (tuned to 10) negative examples for each positive one. 
Results are reported by precision, recall, and F1 for the standard coreference metrics MUC, B\textsuperscript{3}, CEAF-e, and the average F1 of the three metrics, using the official CoNLL scorer \cite{pradhan-etal-2012-conll}.\footnote{\url{https://github.com/conll/reference-coreference-scorers}}

\subsection{Results}
Table~\ref{tab:ecb-results} presents the results on ECB+.
Our model outperforms state-of-the-art results for both the \textsc{Joint} model and the \textsc{Disjoint} event model of \newcite{barhom-etal-2019-revisiting}, with a gain of 1.3 CoNLL $F1$ points and 2.3 CoNLL $F1$ points respectively. The \textsc{Joint} model jointly clusters event and entity mentions, leveraging information across the two subtasks, while the \textsc{Disjoint} event model considers only event mentions, taking the same input as our model. These results assess our model as a suitable baseline for \wecEng{}.

Table~\ref{tab:wec-results} presents the results on \wecEng{}.
First, we observe that despite the certain level of noise in the automatically gathered training data, our model outperforms the same-head-lemma baseline by 9.2 CoNLL $F_1$ points. In fact, it achieves similar error reduction rates relative to the lemma baseline as obtained over ECB+, where training is performed on a clean but smaller training data (18.3\% error reduction in ECB+ and 19.6\% in WEC). 
Furthermore, the performance of both the same-head-lemma baseline and our model are substantially lower on \wecEng{} (Table~\ref{tab:wec-results}) than on ECB+ (Table~\ref{tab:ecb-results}). 
This indicates the more challenging nature of \wecEng{}, possibly due to its corpus wide nature and higher degree of ambiguity (Tables~\ref{tab:table_datasets_stats}).

Further examining the different nature of the two datasets, we applied cross-domain evaluation, applying the ECB+ trained model on \wecEng{} test data and vice versa. The results suggest that due to their different characteristics, with respect to mention type (descriptive vs. referential) and structure (topic-based vs. corpus wide), a model trained on one dataset is less effective (by 8-12 points) when applied to the other (further details are presented in Appendix~\ref{app:cross_domain_evaluation}).

\subsection{Qualitative Analysis}
\label{qualitative_analysis}

To obtain some qualitative insight about the learned models for both ECB+ and \wecEng{}, we manually examined their most certain predictions, looking at the top 5\% instances with highest predicted probability and at the bottom 5\%, of lowest predictions. Some typical examples are given in Appendix~\ref{app:analys}.
Generally, both models tend to assign the highest probabilities to mention pairs that share some lemma, and occasionally to pairs with different lemmas with similar meanings, with the \wecEng{} model making such lexical generalizations somewhat more frequently. 
Oftentimes in these cases, the models fail to distinguish between (gold) positive and negative cases, despite quite clear distinguishing evidence in the context, such as different times or locations. This suggests that the RoBERTa-based modeling of context may not be sufficient, and that more sophisticated models, injecting argument structure more extensively, may be needed.

In both models, the lowest predictions (correctly) correspond mostly to negative mention pairs, and occasionally to positive pairs for which the semantic correspondence is less obvious (e.g. \textit{offered} vs. \textit{candidacy}). 
In addition, we observe that longer spans common in \wecEng{} challenge the span representation model of \citet{lee-etal-2017-end}. This model emphasizes mention boundaries, but these often vary across lexically-similar coreferring mentions with different word order.



\section{Conclusion and Future Work}


In this paper, we presented a generic low-cost methodology and supporting tools for extracting cross-document event coreference datasets from Wikipedia. The methodology was applied to create the larger-scale \wecEng{} corpus, and may be easily applied to additional languages with relatively few adjustments. 
Most importantly, our dataset complements existing resources for the task by 
addressing a different appealing realistic setup: the targeted data is collected across a full corpus rather than within topical document clusters, 
and, accordingly, mentions are mostly referential rather than descriptive.
Hence, we suggest that future research should be evaluated also on \wecEng{}, while future datasets, particularly for other languages, can be created using the WEC methodology and tool suite, all made publicly available.
Our released model provides a suitable baseline for such future work.

\section*{Acknowledgments}

We would like to thank Valentina Pyatkin, Daniela Stepanov and Oren Pereg for their valuable assistance in the data validation process.
The work described herein was supported in part by grants from Intel Labs, Facebook, the Israel Science Foundation grant 1951/17, the Israeli Ministry of Science and Technology and the German Research Foundation through the German-Israeli Project Cooperation (DIP, grant DA 1600/1-1).

\bibliography{wec}

\begin{thebibliography}{26}
\expandafter\ifx\csname natexlab\endcsname\relax\def\natexlab#1{#1}\fi

\bibitem[{Araki et~al.(2014)Araki, Liu, Hovy, and
  Mitamura}]{araki-etal-2014-detecting}
Jun Araki, Zhengzhong Liu, Eduard Hovy, and Teruko Mitamura. 2014.
\newblock \href
  {http://www.lrec-conf.org/proceedings/lrec2014/pdf/963_Paper.pdf} {Detecting
  subevent structure for event coreference resolution}.
\newblock In \emph{Proceedings of the Ninth International Conference on
  Language Resources and Evaluation ({LREC}'14)}, Reykjavik, Iceland. European
  Language Resources Association (ELRA).

\bibitem[{Barhom et~al.(2019)Barhom, Shwartz, Eirew, Bugert, Reimers, and
  Dagan}]{barhom-etal-2019-revisiting}
Shany Barhom, Vered Shwartz, Alon Eirew, Michael Bugert, Nils Reimers, and Ido
  Dagan. 2019.
\newblock \href {https://doi.org/10.18653/v1/P19-1409} {Revisiting joint
  modeling of cross-document entity and event coreference resolution}.
\newblock In \emph{Proceedings of the 57th Annual Meeting of the Association
  for Computational Linguistics}, pages 4179--4189, Florence, Italy.
  Association for Computational Linguistics.

\bibitem[{Bejan and Harabagiu(2010)}]{bejan-harabagiu-2010-unsupervised}
Cosmin Bejan and Sanda Harabagiu. 2010.
\newblock \href {https://www.aclweb.org/anthology/P10-1143} {Unsupervised event
  coreference resolution with rich linguistic features}.
\newblock In \emph{Proceedings of the 48th Annual Meeting of the Association
  for Computational Linguistics}, pages 1412--1422, Uppsala, Sweden.
  Association for Computational Linguistics.

\bibitem[{Chisholm and Hachey(2015)}]{chisholm-hachey-2015-entity}
Andrew Chisholm and Ben Hachey. 2015.
\newblock \href {https://doi.org/10.1162/tacl_a_00129} {Entity disambiguation
  with web links}.
\newblock \emph{Transactions of the Association for Computational Linguistics},
  3:145--156.

\bibitem[{Choubey and Huang(2017)}]{choubey-huang-2017-event}
Prafulla~Kumar Choubey and Ruihong Huang. 2017.
\newblock \href {https://doi.org/10.18653/v1/D17-1226} {Event coreference
  resolution by iteratively unfolding inter-dependencies among events}.
\newblock In \emph{Proceedings of the 2017 Conference on Empirical Methods in
  Natural Language Processing}, pages 2124--2133, Copenhagen, Denmark.
  Association for Computational Linguistics.

\bibitem[{Cohen(1960)}]{doi:10.1177/001316446002000104}
Jacob Cohen. 1960.
\newblock \href {https://doi.org/10.1177/001316446002000104} {A coefficient of
  agreement for nominal scales}.
\newblock \emph{Educational and Psychological Measurement}, 20(1):37--46.

\bibitem[{Cybulska and Vossen(2014)}]{cybulska-vossen-2014-using}
Agata Cybulska and Piek Vossen. 2014.
\newblock Using a sledgehammer to crack a nut? {L}exical diversity and event
  coreference resolution.
\newblock In \emph{Proceedings of the Ninth International Conference on
  Language Resources and Evaluation ({LREC}'14)}, pages 4545--4552, Reykjavik,
  Iceland. European Languages Resources Association (ELRA).

\bibitem[{Dhingra et~al.(2018)Dhingra, Jin, Yang, Cohen, and
  Salakhutdinov}]{dhingra-etal-2018-neural}
Bhuwan Dhingra, Qiao Jin, Zhilin Yang, William Cohen, and Ruslan Salakhutdinov.
  2018.
\newblock \href {https://doi.org/10.18653/v1/N18-2007} {Neural models for
  reasoning over multiple mentions using coreference}.
\newblock In \emph{Proceedings of the 2018 Conference of the North {A}merican
  Chapter of the Association for Computational Linguistics: Human Language
  Technologies, Volume 2 (Short Papers)}, pages 42--48, New Orleans, Louisiana.
  Association for Computational Linguistics.

\bibitem[{Eshel et~al.(2017)Eshel, Cohen, Radinsky, Markovitch, Yamada, and
  Levy}]{eshel-etal-2017-named}
Yotam Eshel, Noam Cohen, Kira Radinsky, Shaul Markovitch, Ikuya Yamada, and
  Omer Levy. 2017.
\newblock \href {https://doi.org/10.18653/v1/K17-1008} {Named entity
  disambiguation for noisy text}.
\newblock In \emph{Proceedings of the 21st Conference on Computational Natural
  Language Learning ({C}o{NLL} 2017)}, pages 58--68, Vancouver, Canada.
  Association for Computational Linguistics.

\bibitem[{Falke et~al.(2017)Falke, Meyer, and
  Gurevych}]{falke-etal-2017-concept}
Tobias Falke, Christian~M. Meyer, and Iryna Gurevych. 2017.
\newblock \href {https://www.aclweb.org/anthology/I17-1081} {Concept-map-based
  multi-document summarization using concept coreference resolution and global
  importance optimization}.
\newblock In \emph{Proceedings of the Eighth International Joint Conference on
  Natural Language Processing (Volume 1: Long Papers)}, pages 801--811, Taipei,
  Taiwan. Asian Federation of Natural Language Processing.

\bibitem[{Honnibal and Montani(2017)}]{spacy2}
Matthew Honnibal and Ines Montani. 2017.
\newblock {spaCy 2}: Natural language understanding with {B}loom embeddings,
  convolutional neural networks and incremental parsing.
\newblock To appear.

\bibitem[{Hovy et~al.(2013)Hovy, Mitamura, Verdejo, Araki, and
  Philpot}]{hovy2013events}
Eduard Hovy, Teruko Mitamura, Felisa Verdejo, Jun Araki, and Andrew Philpot.
  2013.
\newblock Events are not simple: Identity, non-identity, and quasi-identity.
\newblock In \emph{Workshop on events: Definition, detection, coreference, and
  representation}, pages 21--28.

\bibitem[{Joshi et~al.(2019)Joshi, Levy, Zettlemoyer, and
  Weld}]{joshi-etal-2019-bert}
Mandar Joshi, Omer Levy, Luke Zettlemoyer, and Daniel Weld. 2019.
\newblock \href {https://doi.org/10.18653/v1/D19-1588} {{BERT} for coreference
  resolution: Baselines and analysis}.
\newblock In \emph{Proceedings of the 2019 Conference on Empirical Methods in
  Natural Language Processing and the 9th International Joint Conference on
  Natural Language Processing (EMNLP-IJCNLP)}, pages 5802--5807, Hong Kong,
  China. Association for Computational Linguistics.

\bibitem[{Kantor and Globerson(2019)}]{kantor-globerson-2019-coreference}
Ben Kantor and Amir Globerson. 2019.
\newblock \href {https://doi.org/10.18653/v1/P19-1066} {Coreference resolution
  with entity equalization}.
\newblock In \emph{Proceedings of the 57th Annual Meeting of the Association
  for Computational Linguistics}, pages 673--677, Florence, Italy. Association
  for Computational Linguistics.

\bibitem[{Kenyon-Dean et~al.(2018)Kenyon-Dean, Cheung, and
  Precup}]{kenyon-dean-etal-2018-resolving}
Kian Kenyon-Dean, Jackie Chi~Kit Cheung, and Doina Precup. 2018.
\newblock \href {https://doi.org/10.18653/v1/S18-2001} {Resolving event
  coreference with supervised representation learning and clustering-oriented
  regularization}.
\newblock In \emph{Proceedings of the Seventh Joint Conference on Lexical and
  Computational Semantics}, pages 1--10, New Orleans, Louisiana. Association
  for Computational Linguistics.

\bibitem[{Lee et~al.(2017)Lee, He, Lewis, and Zettlemoyer}]{lee-etal-2017-end}
Kenton Lee, Luheng He, Mike Lewis, and Luke Zettlemoyer. 2017.
\newblock \href {https://doi.org/10.18653/v1/D17-1018} {End-to-end neural
  coreference resolution}.
\newblock In \emph{Proceedings of the 2017 Conference on Empirical Methods in
  Natural Language Processing}, pages 188--197, Copenhagen, Denmark.
  Association for Computational Linguistics.

\bibitem[{Liao et~al.(2018)Liao, Lebanoff, and Liu}]{liao-etal-2018-abstract}
Kexin Liao, Logan Lebanoff, and Fei Liu. 2018.
\newblock \href {https://www.aclweb.org/anthology/C18-1101} {{A}bstract
  {M}eaning {R}epresentation for multi-document summarization}.
\newblock In \emph{Proceedings of the 27th International Conference on
  Computational Linguistics}, pages 1178--1190, Santa Fe, New Mexico, USA.
  Association for Computational Linguistics.

\bibitem[{Lin et~al.(2020)Lin, Li, Xin, Li, and Chen}]{lin2020kbpearl}
Xueling Lin, Haoyang Li, Hao Xin, Zijian Li, and Lei Chen. 2020.
\newblock {KBPearl}: {A} {K}nowledge {B}ase {P}opulation {S}ystem {S}upported
  by {J}oint {E}ntity and {R}elation {L}inking.
\newblock \emph{Proceedings of the VLDB Endowment}, 13(7):1035--1049.

\bibitem[{Liu et~al.(2019)Liu, Ott, Goyal, Du, Joshi, Chen, Levy, Lewis,
  Zettlemoyer, and Stoyanov}]{liu2019roberta}
Yinhan Liu, Myle Ott, Naman Goyal, Jingfei Du, Mandar Joshi, Danqi Chen, Omer
  Levy, Mike Lewis, Luke Zettlemoyer, and Veselin Stoyanov. 2019.
\newblock Ro{BERT}a: A robustly optimized {BERT} pretraining approach.
\newblock \emph{arXiv preprint arXiv:1907.11692}.

\bibitem[{Minard et~al.(2016)Minard, Speranza, Urizar, Altuna, van Erp, Schoen,
  and van Son}]{minard-etal-2016-meantime}
Anne-Lyse Minard, Manuela Speranza, Ruben Urizar, Bego{\~n}a Altuna, Marieke
  van Erp, Anneleen Schoen, and Chantal van Son. 2016.
\newblock {MEANTIME}, the {N}ews{R}eader multilingual event and time corpus.
\newblock In \emph{Proceedings of the Tenth International Conference on
  Language Resources and Evaluation ({LREC}'16)}, pages 4417--4422,
  Portoro{\v{z}}, Slovenia. European Language Resources Association (ELRA).

\bibitem[{Nothman et~al.(2012)Nothman, Honnibal, Hachey, and
  Curran}]{nothman2012event}
Joel Nothman, Matthew Honnibal, Ben Hachey, and James~R Curran. 2012.
\newblock Event linking: Grounding event reference in a news archive.
\newblock In \emph{Proceedings of the 50th Annual Meeting of the Association
  for Computational Linguistics (Volume 2: Short Papers)}, pages 228--232.

\bibitem[{Pradhan et~al.(2012)Pradhan, Moschitti, Xue, Uryupina, and
  Zhang}]{pradhan-etal-2012-conll}
Sameer Pradhan, Alessandro Moschitti, Nianwen Xue, Olga Uryupina, and Yuchen
  Zhang. 2012.
\newblock \href {https://www.aclweb.org/anthology/W12-4501} {{C}o{NLL}-2012
  shared task: Modeling multilingual unrestricted coreference in
  {O}nto{N}otes}.
\newblock In \emph{Joint Conference on {EMNLP} and {C}o{NLL} - Shared Task},
  pages 1--40, Jeju Island, Korea. Association for Computational Linguistics.

\bibitem[{Singh et~al.(2012)Singh, Subramanya, Pereira, and
  McCallum}]{singh12:wiki-links}
Sameer Singh, Amarnag Subramanya, Fernando Pereira, and Andrew McCallum. 2012.
\newblock Wikilinks: A large-scale cross-document coreference corpus labeled
  via links to {Wikipedia}.
\newblock Technical Report UM-CS-2012-015.

\bibitem[{Upadhyay et~al.(2016)Upadhyay, Gupta, Christodoulopoulos, and
  Roth}]{upadhyay-etal-2016-revisiting}
Shyam Upadhyay, Nitish Gupta, Christos Christodoulopoulos, and Dan Roth. 2016.
\newblock \href {https://www.aclweb.org/anthology/C16-1183} {Revisiting the
  evaluation for cross document event coreference}.
\newblock In \emph{Proceedings of {COLING} 2016, the 26th International
  Conference on Computational Linguistics: Technical Papers}, pages 1949--1958,
  Osaka, Japan. The COLING 2016 Organizing Committee.

\bibitem[{Vossen et~al.(2018)Vossen, Ilievski, Postma, and
  Segers}]{vossen-etal-2018-dont}
Piek Vossen, Filip Ilievski, Marten Postma, and Roxane Segers. 2018.
\newblock \href {https://www.aclweb.org/anthology/L18-1480} {Don{'}t annotate,
  but validate: a data-to-text method for capturing event data}.
\newblock In \emph{Proceedings of the Eleventh International Conference on
  Language Resources and Evaluation ({LREC} 2018)}, Miyazaki, Japan. European
  Language Resources Association (ELRA).

\bibitem[{Wang et~al.(2019)Wang, Yu, Guo, Das, Xiong, and
  Gao}]{wang-etal-2019-multi-hop}
Haoyu Wang, Mo~Yu, Xiaoxiao Guo, Rajarshi Das, Wenhan Xiong, and Tian Gao.
  2019.
\newblock \href {https://doi.org/10.18653/v1/D19-5813} {Do multi-hop readers
  dream of reasoning chains?}
\newblock In \emph{Proceedings of the 2nd Workshop on Machine Reading for
  Question Answering}, pages 91--97, Hong Kong, China. Association for
  Computational Linguistics.

\end{thebibliography}
\bibliographystyle{acl_natbib}

\clearpage
\appendix

\section{WEC Dataset}

\subsection{Infoboxs Distillation}
\label{app:subevent}


Excluding infobox types of broad general events, consisting of many sub-events is necessary as often Wikipedia authors link to a broad event from anchor texts that refer to a subevent (and should be regarded as non-coreferring by standard definitions of the event coreference task \cite{hovy2013events,araki-etal-2014-detecting}). For example, in English Wikipedia many event articles containing the infobox \emph{``Election"} tend to be pointed at from anchor texts that describe subevents, such as \textit{2016 primaries} linking to \textit{2016 United States presidential election}.
Additionally, for some infobox types, pointing mentions often correspond to related, but not coreferring, named entities; hence, we discard the infobox types to avoid such noisy mentions. For example, articles of the infobox type ``Race" are often linked from mentions denoting the name of the country in which the race took place. Table~\ref{tab:disq_infoboxs} presents excluded infobox types due to the above reasons.

\subsection{WEC-Eng Infobox Types}
\label{app:infobox}

As mentioned in the paper~(Section~\ref{sec:wec}), we manually explored the various Wikipedia infobox types and selected only those denoting an event. Table~\ref{tab:table_infoboxs} shows the number of coreference clusters (event pages) and mentions for each selected infobox type. In the table, infobox types falling under the same general category are grouped together.

\subsection{Lexical Distribution of ECB+ and \wecEng{} Mentions}
\label{app:verb_dist}

Event mentions can appear in various lexical forms in a document, such as verbs (e.g. \textit{exploded}), nominalization (e.g. \textit{crash}), common nouns (e.g. \emph{party, accident}) and proper nouns (e.g. \textit{Cannes Festival 2016}).
In order to have a rough estimation of the distribution of these different forms, we manually analyze 100 sampled mentions, from each of \wecEng{} and ECB+, and present the statistics in Table~\ref{tab:verbs}.

\section{Further Analysis}

\subsection{Cross-Domain Experiment and Results}
\label{app:cross_domain_evaluation}

To further assess the difference between the ECB+ and \wecEng{} datasets, we evaluate the cross-domain ability of the model trained on one dataset, and tested on the other one. We use the same pairwise model as in Tables~\ref{tab:ecb-results} and~\ref{tab:wec-results}, while only tuning the stop criterion for the agglomerative clustering, in the corresponding development set. 
Table~\ref{tab:results_cross_domain} presents the cross-domain performance of our model (1) trained on ECB+ and evaluated on \wecEng{} and (2) trained on \wecEng{} and evaluated on ECB+, together with the corresponding in-domain performances (from Tables~\ref{tab:ecb-results} and~\ref{tab:wec-results}) for comparison.
The performance drop for both cross-domain experiments indicates that these two datasets address different challenges of cross-document event coreference resolution. 

\subsection{Qualitative Analysis Examples}
\label{app:analys}

Illustrating Section~\ref{qualitative_analysis}, Tables ~\ref{tab:ecb_pair_analys} and~\ref{tab:wec_pair_analys} present examples sampled from the predictions of the ECB+ and \wecEng{} models on their respective development sets.


\begin{table*}
    \centering
    \resizebox{1\textwidth}{!}{
    \begin{tabular}{@{}c|l|ll@{}}
    \toprule
    \textbf{Infobox} & \multicolumn{1}{c}{\textbf{Mention link and context}} & \multicolumn{2}{c}{\textbf{Non-coreferring relation}}  \\
    \toprule
        \multirow{3}{*}{Race} & ..He crashed in \emph{\textcolor{blue}{Italy}}, and finished two laps off the lead in Portugal.. & \textbf{\textcolor{red}{\xmark}} & Location \\
        & ..after several unrewarded drives the year before, namely in \emph{\textcolor{blue}{Italy}}.. & \textbf{\textcolor{red}{\xmark}} & Location \\
        & ..did not appear at the \emph{\textcolor{blue}{British}} or German Grands Prix.. & \textbf{\textcolor{red}{\xmark}} & Location \\
        \midrule
        \multirow{2}{*}{Sport} & ..in the \emph{\textcolor{blue}{Gold Medal final}} straight sets by an identical score of 16 – 21.. & \textbf{\textcolor{red}{\xmark}} & Subevent \\
        & ..2012 FINA World Swimming Championships \emph{\textcolor{blue}{(25 m)}} and in the 50 m freestyle.. & \textbf{\textcolor{red}{\xmark}} & Subevent  \\
        \midrule
        \multirow{3}{*}{Wars} & ..throughout most of World War II, in France \emph{\textcolor{blue}{Holland}} and Belgium.. & \textbf{\textcolor{red}{\xmark}} & Location \\
        & ..During the \emph{\textcolor{blue}{ensuing campaign}} the Netherlands were defeated and occupied .. & \textbf{\textcolor{red}{\xmark}} & Subevent  \\ & ..and made possible further \emph{\textcolor{blue}{advance into Crimea}} and industrially developed Eastern.. & \textbf{\textcolor{red}{\xmark}} & Subevent \\
    \bottomrule
    \end{tabular}}
    \caption{Sample of disqualified general infoboxs types, usually linked from many erroneous mentions}
    \label{tab:disq_infoboxs}
\end{table*}
\begin{table*}[!t]
    \centering
    \resizebox{0.5\textwidth}{!}{
    \begin{tabular}{@{}lrr@{}}
    \toprule
    \textbf{Infobox Type} & \textbf{Event Pages} & \textbf{Mentions} \\
    \midrule
    Awards (2 Infoboxes) & 1,848 & 8,910 \\
    Meetings (4 Infoboxes) & 1,212 & 5,942 \\
    Civilian Attack & 1,178 & 9,490 \\
    Airliner Accident (6 Infoboxes) & 596 & 2,640 \\
    Festival & 541 & 2,770 \\
    Beauty Pageant & 431 & 1,616 \\
    Earthquake & 387 & 3,526 \\
    Contest & 268 & 2,029 \\
    Concert & 262 & 1,082 \\
    News Event & 241 & 1,663 \\
    Terrorist Attack & 230 & 1,296 \\
    Wildfire & 135 & 720 \\
    Flood & 121 & 990 \\
    Weapons Test (2 Infoboxes) & 65 & 345 \\
    Eruption & 31 & 343 \\
    Solar Eclipse & 31 & 121 \\
    Oilspill & 18 & 184 \\
    Rail Accident & 2 & 5 \\
    \midrule
    Total (28) & 7,597 & 43,672 \\
    \bottomrule
    \end{tabular}}
    \caption{Final 28 \wecEng{} extracted infobox types list}
    \label{tab:table_infoboxs}
\end{table*}

\begin{table*}
    \centering
    \resizebox{.58\textwidth}{!}{
    \begin{tabular}{@{}llcc@{}}
        \toprule
        & \phantom{aaahha} & \makecell{\wecEng{}} & \makecell{ECB+} \\
        \midrule
        \# Verbs && 7 & 62 \\
        \# Proper nouns && 38 & 2 \\
        \# Nominalizations && 24 & 23 \\
        \# Other common nouns && 31 & 13 \\
        \bottomrule
    \end{tabular}}
    \caption{Distribution of lexical types for 100 randomly sampled mentions from ECB+ and \wecEng{}}
    \label{tab:verbs}
\end{table*}

\begin{table*}
    \centering
    \resizebox{\textwidth}{!}{
    \begin{tabular}{@{}lcccccccccccccc@{}}\toprule
    & \phantom{abc}&\multicolumn{3}{c}{MUC} & \phantom{abc}& \multicolumn{3}{c}{$B^3$} & \phantom{abc}& \multicolumn{3}{c}{$CEAF$} & \phantom{abc}& CoNLL\\
    \cmidrule{3-5} \cmidrule{7-9} \cmidrule{11-13} \cmidrule{15-15}
    && R & P & $F_1$ && R & P & $F_1$ && R &P & $F_1$ && $F_1$  \\ 
    \midrule
    \wecEng{} to \wecEng{} && 78 & 83.6 & 80.7 && 66.1 & 55.3 & 60.2 && 53.4 & 40.3 & 45.9 && \textbf{62.3} \\
       ECB+ to \wecEng{} && 70.1 & 79.3 & 74.4 && 53.8 & 54.6 &  54.2 && 42.9 & 27.4 & 33.5 && 54 \\
       \midrule
       ECB+ to ECB+ && 84.2 & 81.8 & 83 && 80.8 & 81.7 & 81.3 && 76.7 & 79.5 & 78 && \textbf{80.8} \\
       \wecEng{} to ECB+ && 77.1 & 68.5 & 72.6 && 74.2 & 73.1 & 73.6 && 56.4 & 66.5 & 61 && 69.1 \\
    \bottomrule
    \end{tabular}}
    \caption{Cross-domain results of model trained on ECB+ and evaluated on \wecEng{} test set, and vice-versa}
    \label{tab:results_cross_domain}
\end{table*}


\begin{table*}
    \centering
    \resizebox{.95\textwidth}{!}{
    \begin{tabular}{@{}c|l@{}}
    \toprule
    \textbf{Gold} & \multicolumn{1}{c}{\textbf{Mention Link and Context}} \\
    \toprule
        \multirow{1}{*}{} & \multicolumn{1}{c}{Highest Predicted Probabilities} \\
        \midrule
        \multirow{2}{*}{+} & ..Nov 12 2011, 01:19 hrs Warship INS Sukanya on Thursday foiled a \emph{\textcolor{blue}{ piracy}} attempt in the Gulf of Aden.. \\ & ..November 11, 2011 18:50 IST Indian Naval ship, INS Sukanya, thwarted a \emph{\textcolor{blue}{ piracy}} attack in the Gulf of Aden and.. \\
        \midrule
        \multirow{2}{*}{+} & ..We are thrilled to have Ellen DeGeneres \emph{\textcolor{blue}{ host}} the Oscars, "producers Craig Zadan and Neil Meron said in a.. \\ & ..the Oscars Published August 02 , 2013 Ellen DeGeneres will \emph{\textcolor{blue}{ host}} the Oscars for a second time next year.. \\
        \midrule
        \multirow{2}{*}{+} & ..this morning that Ellen DeGeneres would return to the \emph{\textcolor{blue}{ Academy Awards}} as emcee , marking her second hosting stint.. \\ & ..DeGeneres just tweeted she will be hosting the \emph{\textcolor{blue}{Oscars}} this year.. \\
        \Xhline{2\arrayrulewidth}
        \multirow{2}{*}{-} & ..A British man has been killed after \emph{\textcolor{blue}{ falling}} around 2,000ft.. \\ & ..A 36-year-old Australian climber is dead after \emph{\textcolor{blue}{ falling}} about 150 metres.. \\
        \midrule
        \multirow{2}{*}{-} & ..Comedian Ellen DeGeneres picked to host 2014 \emph{\textcolor{blue}{ Oscars}} Fri Aug 2, 2013 6:50pm EDT.. \\ & ..DeGeneres , the star of her own daytime talk show "Ellen," first hosted the \emph{\textcolor{blue}{ Oscars}} in 2007, becoming only.. \\
        \midrule
        \multirow{2}{*}{-} & ..play in Sunday 's AFC divisional \emph{\textcolor{blue}{ playoff}} game at Pittsburgh.. \\ & ..drink-driving five days before a key NFL \emph{\textcolor{blue}{ playoff}} game.. \\
        \midrule
        \multirow{1}{*}{} & \multicolumn{1}{c}{Lowest Predicted Probabilities} \\
        \midrule
        \multirow{2}{*}{-} & ..I'm hosting the Oscars! "DeGeneres \emph{\textcolor{blue}{ tweeted}} Friday.. \\ & ..The comedian and talk show host made the \emph{\textcolor{blue}{ announcement}} via her Twitter feed.. \\
        \midrule
        \multirow{2}{*}{-} & ..and chief medical correspondent for CNN, has been \emph{\textcolor{blue}{ offered}} the post of Surgeon General.. \\ & ..Since CNN learned of Gupta's \emph{\textcolor{blue}{ candidacy}}, the cable network.. \\
        \midrule
        \multirow{2}{*}{-} & ..Cheeks returns as Sixers' fixer Jim O'Brien was \emph{\textcolor{blue}{ shown the door}} after a single tumultuous season.. \\ & ..The Sixers made a power move this offseason, \emph{\textcolor{blue}{ firing}} head coach Jim O'Brien.. \\
        \midrule
        \multirow{2}{*}{-} & ..the California Highway Patrol \emph{\textcolor{blue}{ said}}. Williams, 32, was going west in his 2008 Bentley.. \\ & ..\emph{\textcolor{blue}{ According to}} the California Highway Patrol, defensive tackle Jamal Williams.. \\
    \bottomrule
    \end{tabular}}
    \caption{Examples of ECB+ highest predicted probabilities and the lowest predicted probabilities, along with their gold standard label.}
    \label{tab:ecb_pair_analys}
\end{table*}

\begin{table*}
    \centering
    \resizebox{.95\textwidth}{!}{
    \begin{tabular}{@{}c|l@{}}
    \toprule
    \textbf{Gold} & \multicolumn{1}{c}{\textbf{Mention Link and Context}} \\
    \toprule
        \multirow{1}{*}{} & \multicolumn{1}{c}{Highest Predicted Probabilities} \\
        \midrule
        \multirow{2}{*}{+} & ..Bangladesh Army He was the chief organiser of the \emph{\textcolor{blue}{ assassination}} of Sheikh Mujibur Rahman.. \\ & ..When Sheikh Mujibur Rahman was \emph{\textcolor{blue}{ killed}} on 15 August 1975 by members of the Bangladesh Army.. \\
        \midrule
        \multirow{2}{*}{+} & ..On 6 April 1968 , the Hall was the host venue for the \emph{\textcolor{blue}{Eurovision Song Contest}} which was.. \\
        & ..that at the \emph{\textcolor{blue}{1968 Contest}}, the voting had been rigged by Spanish dictator Francisco Franco and.. \\
        \midrule
        \multirow{2}{*}{+} & ..Dagestani separatist movement, combined with a series of \emph{\textcolor{blue}{apartment bombings}} in Russia.. \\ & ..rebellious Chechnya republic as a retaliation for \emph{\textcolor{blue}{terrorist bombings}} in Moscow and other cities.. \\
        \Xhline{2\arrayrulewidth}
        \multirow{2}{*}{-} & ..towns and villages were destroyed by the \emph{\textcolor{blue}{ earthquake}} and alluvium on May 31, 1970.. \\ & ..passing through the area struck by an \emph{\textcolor{blue}{ earthquake}} in May 2012.. \\ 
        \midrule 
        \multirow{2}{*}{-} & ..sent troops to the rebellious Chechnya republic as a retaliation for \emph{\textcolor{blue}{terrorist bombings}} in Moscow and.. \\ & ..Alaoui died from injuries suffered in a \emph{\textcolor{blue}{terrorist attack}} in Ouagadougou, Burkina Fas.. \\
        \midrule
        \multirow{2}{*}{-} & ..in monasteries in Austria-Hungary, after the recent \emph{\textcolor{blue}{ massacres}} and expulsions of Albanians by their Slavic neighbours.. \\ & ..gained international fame in 2014 through the \emph{\textcolor{blue}{ genocide}} of the Islamic State on the Yazidis.. \\ 
        \midrule
        \multirow{2}{*}{-} & ..and was a gold Democrat delegate to the \emph{\textcolor{blue}{ 1896 Democratic National Convention}}.. \\ & ..while watching Pat Buchanan's televised speech at the \emph{\textcolor{blue}{ 1996 Republican National Convention}}.. \\
        \midrule
        \multirow{1}{*}{} & \multicolumn{1}{c}{Lowest Predicted Probabilities} \\
        \midrule
        \multirow{2}{*}{-} & ..As the official representative of her country to the 2011 \emph{\textcolor{blue}{ Miss Universe}} pageant.. \\ & ..as the final was held too late to send the winner to the \emph{\textcolor{blue}{2011 edition}}.. \\
        \midrule
        \multirow{2}{*}{-} & ..the ground the funeral party came under a hand-grenade \emph{\textcolor{blue}{ attack}} from a lone Loyalist paramilitary.. \\ & ..the Hill bar shooting, the \emph{\textcolor{blue}{ Milltown massacre}}, the Sean Graham's and James Murray's bookmakers' shootings.. \\
        \midrule
        \multirow{2}{*}{-} & ..Christian-Jaque's 1946 film ''A Lover's Return'' was entered into the \emph{\textcolor{blue}{ 1946 Cannes Film Festival}}.. \\ & ..This film won two prizes at the first \emph{\textcolor{blue}{ Cannes Film Festival in 1946}}.. \\
        \midrule
        \multirow{2}{*}{-} & ..The epicenter of the \emph{\textcolor{blue}{ 2006 Kiholo Bay earthquake}} was some offshore of the village.. \\ & ..However, the unit was damaged in an \emph{\textcolor{blue}{ earthquake in October 2006}}.. \\
    \bottomrule
    \end{tabular}}
    \caption{Examples of \wecEng{} highest predicted Probabilities and the lowest predicted probabilities, along with their gold standard label.}
    \label{tab:wec_pair_analys}
\end{table*}



\end{document}